\documentclass[journal]{IEEEtran}

\usepackage{cite}
\usepackage{graphicx}
\usepackage{multirow}
\usepackage[pagebackref=true,breaklinks=true,letterpaper=true,colorlinks,bookmarks=false]{hyperref}
\usepackage{amsmath}
\usepackage{amssymb}
\usepackage{color}

\usepackage{pgfplots}
\pgfplotsset{width=9cm, compat=1.9}

\newcommand{\etal}{\textit{et al.}\@ }
\newcommand{\ie}{\textit{i.e.}\@ }

\begin{document}

\title{A Novel Generator with Auxiliary Branch \\ for Improving GAN Performance
}




\author{Seung Park and Yong-Goo~Shin,~\IEEEmembership{Member}, IEEE

\thanks{S. Park is with Biomedical Engineering, Chungbuk National University Hospital, Chungbuk National University College of Medicine, Cheongju-si, Chungcheongbuk-do, 28644, Rep. of Korea (e-mail: spark.cbnuh@gmail.com, spark90@chungbuk.ac.kr). Y.-G. Shin is with the Department of Electronics and Information Engineering, Korea University, Sejong-si, Rep. of Korea (e-mail: ygshin92@korea.ac.kr, corresponding author)}
\thanks{This work was supported by the National Research Foundation of Korea (NRF) grant funded by the Korea government (MSIT) (No.2022R1G1A1004001). This research was supported by the MSIT (Ministry of Science and ICT), Korea under the ITRC(Information Technology Research Center) support program (IITP-2023-RS-2023-00258971) supervised by the IITP (Institute for Information \& Communications Technology Planning \& Evaluation).}
}


\markboth{IEEE transactions on neural networks and learning systems (Accept, To be published)}%
{Shell \MakeLowercase{\textit{Park et al.}}}

\maketitle

\begin{abstract}
The generator in the generative adversarial network (GAN) learns image generation in a coarse-to-fine manner in which earlier layers learn the overall structure of the image and the latter ones refine the details. To propagate the coarse information well, recent works usually build their generators by stacking up multiple residual blocks. Although the residual block can produce a high-quality image as well as be trained stably, it often impedes the information flow in the network. To alleviate this problem, this brief introduces a novel generator architecture that produces the image by combining features obtained through two different branches: the main and auxiliary branches. The goal of the main branch is to produce the image by passing through the multiple residual blocks, whereas the auxiliary branch is to convey the coarse information in the earlier layer to the later one. To combine the features in the main and auxiliary branches successfully, we also propose a gated feature fusion module that controls the information flow in those branches. To prove the superiority of the proposed method, this brief provides extensive experiments using various standard datasets including CIFAR-10, CIFAR-100, LSUN, CelebA-HQ, AFHQ, and tiny-ImageNet. Furthermore, we conducted various ablation studies to demonstrate the generalization ability of the proposed method. Quantitative evaluations prove that the proposed method exhibits impressive GAN performance in terms of Inception score (IS) and Frechet inception distance (FID). For instance, the proposed method boosts the FID and IS scores on the tiny-ImageNet dataset from 35.13 to 25.00 and 20.23 to 25.57, respectively.  


\end{abstract}

\section{Introduction}
\label{sec1}
\IEEEPARstart{G}{enerative} adversarial network (GAN)~\cite{goodfellow2014generative}, is an algorithmic framework that synthesizes a target data distribution from a given prior distribution. GAN has attracted great attention by achieving great success for various applications including text-to-image translation~\cite{reed2016generative, hong2018inferring}, image-to-image translation~\cite{isola2017image, choi2018stargan, zhu2017unpaired}, and image inpainting~\cite{sagong2019pepsi, shin2020pepsi++}. In general, GAN is composed of two different networks called generator and discriminator. The goal of the generator is to produce the real data distribution to fool the discriminator, whereas the discriminator is trained to classify the real sample from the fake one which is synthesized by the generator. Since the goals of the generator and discriminator are opposed to each other, GAN is difficult to train stably compared with the supervised learning-based convolutional neural networks (CNN). 

Recent studies observed that the sharp gradient space of the discriminator is the major reason for the unstable problem in the GAN training~\cite{wu2021gradient}. To resolve this problem, some papers~\cite{miyato2018spectral, gulrajani2017improved, zhang2019consistency} introduced regularization or normalization techniques that inhibit the discriminator from making the sharp gradient space. As the regularization method, the conventional methods~\cite{gulrajani2017improved, wu2019generalization, wei2018improving} usually added the regularization term into the adversarial loss function. Among the various regularization approaches, gradient penalty-based regularization methods~\cite{gulrajani2017improved, wu2019generalization, wei2018improving}, which add the gradient norm as the penalty term, are the most popular techniques and have been widely used. The normalization methods~~\cite{miyato2018spectral, arjovsky2017wasserstein, kurach2019large} also have been widely studied. Spectral normalization (SN)~\cite{miyato2018spectral} is the most popular one that controls the Lipschitz constraint on the discriminator by constraining the spectral norm of each layer. Since these regularization and normalization methods can lead the stable GAN training, recent studies usually employ these techniques when training their networks. 

Some researchers~\cite{park2021GRB, miyato2018cgans, zhang2019self, yeo2021simple} studied architectural modules to improve the GAN performance. Miyato~\etal~\cite{miyato2018cgans} introduced a conditional projection technique that provides conditional information to the discriminator by projecting the conditional weight vector to the feature vector. This method significantly boosts the performance of the conditional GAN (cGAN) scheme. Zhang~\etal~\cite{zhang2019self} introduced a self-attention module that guides the generator and discriminator on where to attend. Yeo~\etal~\cite{yeo2021simple} introduced a simple yet effective technique, called a cascading rejection module, which produces dynamic features in an iterative manner at the last layer of the discriminator. Li~\etal~\cite{li2023dw} proposed a dynamic weighted GAN (DW-GAN) having multiple discriminators, which synthesizes the images following the color tone of the target image. These methods generate images with plausible quality but still use the common generator architecture. 

On the other hand, some papers investigated new forms of convolution layer to improve the GAN performance. Park~\etal~\cite{park2021generative} proposed a perturbed convolution that prevents the discriminator from falling into the overfitting problem. However, since PConv is developed for the discriminator, it is hard to apply to the generator. Sagong~\etal~\cite{sagong2019cgans} introduced a conditional convolution layer for the generator, which incorporates the conditional information into the convolution operation. Although this method shows fine performance on cGAN, it has a limitation: this method only can be used for cGAN scheme since it is designed to replace the conditional batch normalization. Recently, Park~\etal~\cite{park2021generative_conv} proposed a novel convolution layer, called a generative convolution, which controls the convolution weight of the generator according to the given latent vector. 

There have been several attempts to newly design the generator and discriminator architectures for boosting the GAN performance. Specifically, in the early stage of the GAN study, Radford~\etal~\cite{radford2015unsupervised} designed the generator and discriminator architectures by stacking the transposed convolution layer and standard convolution layer in series, respectively. These network architectures are simple to build but generate low-quality images. Karras~\etal~\cite{karras2017progressive} introduced the GAN with a novel training strategy, called a progressive growing skill, which effectively leads the generator and discriminator to produce high-resolution images. Recently, Miyato~\etal~\cite{miyato2018cgans, miyato2018spectral} proposed a spectrally normalized GAN~(SNGAN) that builds generator and discriminator by stacking multiple residual blocks~\cite{he2016deep}. To further improve the SNGAN performance, Brock~\etal~\cite{brock2018large} proposed a new generative model, called BigGAN, which has a SNGAN-like structure in which noise is additionally input to the residual block. Since SNGAN and BigGAN are easy to implement and train stably, most recent studies employ these networks as their baseline networks. However, even though SNGAN and BigGAN exhibit fine GAN performance, we believe that designing a novel generator and discriminator architectures for improving the GAN performance still remains an open question. Therefore, this brief mainly focuses on developing the novel generator architecture. 

A common characteristic of generators in SNGAN and BigGAN is that they are built by stacking multiple residual blocks in series. These generators produce images in a coarse-to-fine manner, where the earlier blocks provide a rough template and the latter ones clarify the details. However, as pointed out in~\cite{huang2017densely}, even if the residual block is used, the information flow in the generator may be impeded. That means it may be difficult to impart coarse information in the earlier blocks to the latter ones. To alleviate this issue, this brief proposes a novel generator architecture that embraces the strength point of the conventional methods as well as further improves the GAN performance. The proposed method consists of two different branches: the main and auxiliary branches. The main branch learns features by passing through the multiple residual blocks, whereas the auxiliary branch propagates the coarse information in the earlier residual block to the later one. In other words, the goal of the auxiliary branch is to assist the information flow in the generator. In order to combine the features in both branches effectively, we introduce a gated feature fusion module that controls the information flow in those branches. To demonstrate the superiority of the proposed method, this brief presents extensive experimental results using various standard datasets such as CIFAR-10~\cite{krizhevsky2009learning}, CIFAR-100~\cite{krizhevsky2009learning}, LSUN~\cite{yu15lsun}, CelebA-HQ~\cite{karras2017progressive}, AFHQ~\cite{choi2020stargan},  and tiny-ImageNet~\cite{23deng2009imagenet, yao2015tiny}. Furthermore, we provided various ablation studies to demonstrate the generalization ability of the proposed method. Quantitative evaluations show that the proposed method significantly improves GAN performance in terms of Inception score (IS) and Frechet inception distance (FID).  

Key contributions of our brief are as follows: First, we propose a novel generator architecture that contains the auxiliary branch and gated feature fusion module, which support the information flow in the generator. Second, with slight additional hardware costs, the proposed method significantly improves the GAN and conditional GAN performances on various datasets in terms of FID and IS scores. For instance, the proposed method improves FID and IS scores on the tiny-ImageNet dataset from 35.13 to 25.00 and 20.23 to 25.57, respectively.

\section{Background}
\label{sec2}
\subsection{Generative Adversarial Network}
\label{sec2.1}
GAN~\cite{goodfellow2014generative} is a min-max game between two different networks called generator $G$ and discriminator $D$. $G$ is trained to generate visually pleasing images, whereas $D$ is optimized to classify the generated images from real ones. However, both networks compete with each other, GAN is hard to train stably compared with supervised learning. To address this problem, some studies~\cite{mao2017least, arjovsky2017wasserstein, lim2017geometric, gulrajani2017improved} have been conducted to reformulate the objective function. For instance, Mao~\etal~\cite{mao2017least} introduced the objective function using the least square errors, whereas Arjovsky~\etal~\cite{arjovsky2017wasserstein} proposed Wasserstein GAN~(WGAN) that computes the Wasserstein distance between the real and generated images. Another notable objective function is a hinge-adversarial loss~\cite{lim2017geometric}, which is widely used in practice~\cite{yeo2021simple, miyato2018cgans, miyato2018spectral, brock2018large, park2021generative, sagong2019pepsi, shin2020pepsi++, park2021generative_conv, park2021GRB}. On the other hand, cGAN, which produces class-conditional images, has also been actively studied~\cite{mirza2014conditional, odena2017conditional, miyato2018cgans}. In general, cGAN utilizes additional conditional information such as class labels or text conditions, in order to produce the condition-specific feature. By optimizing the objective functions for cGAN~\cite{mirza2014conditional, miyato2018cgans}, the generator can select the image class to be generated.

\subsection{Residual Block-based Generator}
\label{sec2.2}
\begin{figure}
\centering
\includegraphics[width=0.99\linewidth]{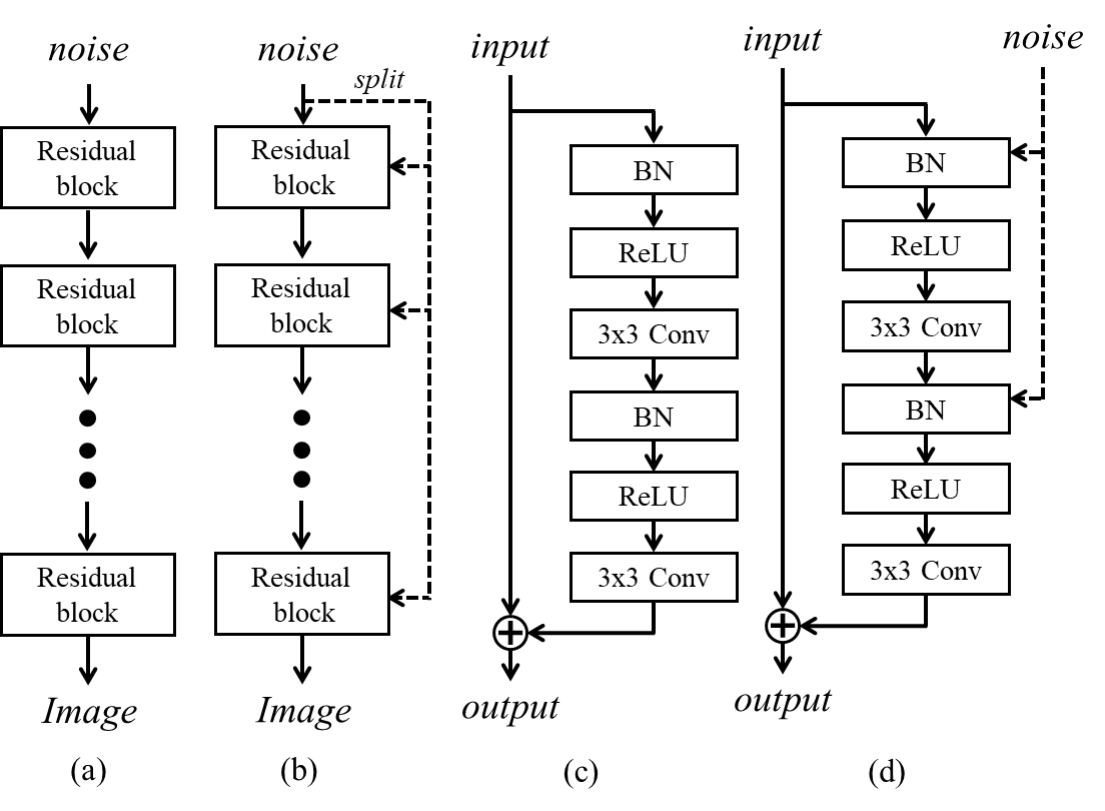}
\caption{The overall generator architectures and residual blocks utilized in SNGAN~\cite{miyato2018spectral, miyato2018cgans} and BigGAN~\cite{brock2018large}. (a) Generator architecture for SNGAN, (b) Generator architecture for BigGAN, (c) Residual block for SNGAN, (d) Residual block for BigGAN.}
\label{fig:fig1}
\vspace{-0cm}
\end{figure}

In the field of GAN-based image generation, SNGAN~\cite{miyato2018cgans, miyato2018spectral} and BigGAN~\cite{brock2018large} are the most popular base networks since they are simple and exhibit fine performance. The overall generator architectures of those methods are depicted in Fig.~\ref{fig:fig1}. More specifically, in the SNGAN paper, they built the generator using the multiple standard residual blocks. To further improve the GAN performance, Brock~\etal~\cite{brock2018large} introduced a residual block in which the noise vector is additionally input to infer the scaling and shifting parameters of the batch normalization layer. This approach slightly modifies the residual block but still has a similar generator architecture to SNGAN. Indeed, as observed in the previous studies~\cite{karras2019style, yang2021semantic}, the generator produces images in a coarse-to-fine manner, where the earlier layers provide a rough template and the latter layers refine the details. That means it needs to convey the information in the earlier layers to the latter ones well. However, since the generators in SNGAN and BigGAN only contain the multiple residual blocks, it may impede the information flow in the network~\cite{huang2017densely}. 

To validate of our hypothesis, we conducted simple experiments: we replaced the residual block with the dense block~\cite{huang2017densely} which propagates the previous information to the latter much better. In our experiments, we used BigGAN as the baseline model and evaluated the GAN performance in terms of IS and FID on the CIFAR-10~\cite{krizhevsky2009learning} and CIFAR-100~\cite{krizhevsky2009learning} datasets. The detailed descriptions of the network architectures and evaluation metrics will be explained in Section~\ref{subsec:3.2} and~\ref{subsec:4.1}, respectively. As shown in Table~\ref{table1} summarizing the experimental results, when replacing the residual block with the dense block, the GAN performances are improved on both datasets. In addition, to match the number of network parameters, we conducted ablation studies that increase the channel dimensions of the residual block (Residual* in Table~\ref{table1}). Although Residual* shows better performance than the residual block, it is still weak compared with the dense block. That means even with the same number of network parameters, the dense block is more effective to achieve fine GAN performance than the residual block. Based on these results, we expect that the GAN performance could be improved by propagating the information in the earlier layers to the latter ones well.

\section{Proposed Method}
\label{sec3}
\subsection{Generator with Auxiliary Branch}
\label{subsec3.1}
\begin{figure}
\centering
\includegraphics[width=0.85\linewidth]{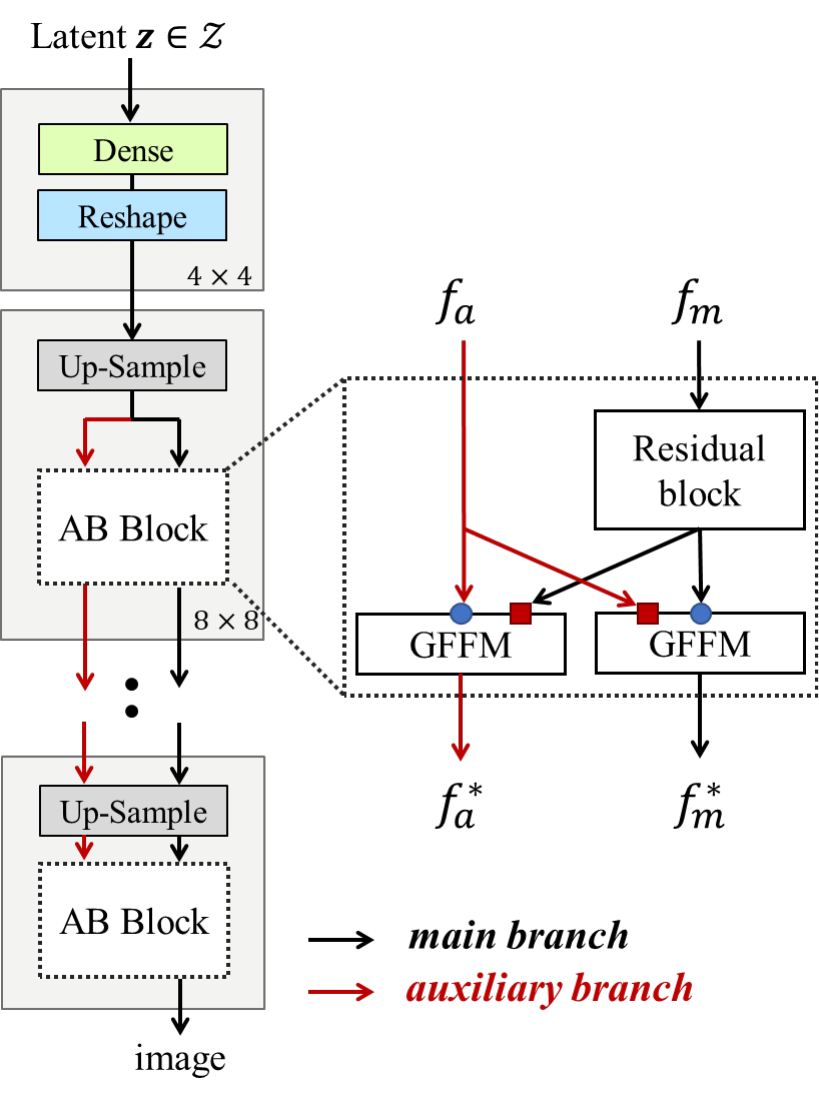}
\caption{The overall generator architecture of the proposed method. The proposed method builds the generator using multiple AB blocks which contain the main and auxiliary branches.}
\label{fig:fig2}
\vspace{-0.5cm}
\end{figure}

This brief introduces the novel generator architecture including main and auxiliary branches. Fig.~\ref{fig:fig2} illustrates the layout of the proposed method schematically. The goal of the auxiliary branch is to convey the coarse information in the earlier layers to the latter ones. Indeed, an intuitive way to flow information well is to directly sum or concatenate features in the earlier layers to the later ones. However, we should notice that the feature levels in earlier and later layers are different. More specifically, the earlier layers produce high-dimensional features that contain coarse information, whereas the latter ones learn low-dimensional features for image details. Therefore, it may not be the best choice to directly sum or concatenate those features. In addition, the latter layers do not need all information in the earlier layers; they may only need some parts of the information in the earlier layers. That means some procedures are required to select the appropriate features and combine them successfully. 

\begin{table}[t]
\caption{Comparison of the dense block with residual one on the CIFAR-10 and CIFAR-100 datasets in terms of IS and FID. The best results are in bold.}
\begin{center}
\begin{tabular}{c | c | c | c | c | c | c | c}
\hline
\multicolumn{2}{c|}{Block name} & \multicolumn{2}{c|}{Residual} & \multicolumn{2}{c|}{Residual*} & \multicolumn{2}{c}{Dense} \\
\hline
\multicolumn{2}{c|}{Parameters} & \multicolumn{2}{c|}{3.78M} & \multicolumn{2}{c|}{4.74M} & \multicolumn{2}{c}{4.74M}\\ 
\hline
Dataset & & IS$\uparrow$ & FID$\downarrow$ & IS$\uparrow$ & FID$\downarrow$ & IS$\uparrow$ & FID$\downarrow$ \\
\hline
\multirow{5}*{CIFAR-10} & trial 1 & 7.70 & 13.98 & 7.74 & 12.93 & 7.96 & 12.41 \\
 & trial 2 & 7.80 & 13.69 & 7.68 & 13.34 & 7.95 & 12.20 \\
 & trial 3 & 7.87 & 13.09 & 7.79 & 12.93 & 7.92 & 12.04 \\
\cline{2-8}
 & Mean & 7.79 & 13.58 & 7.74 & 13.07 & \textbf{7.94} & \textbf{12.22} \\
\cline{2-8} 
 & Std & 0.08 & 0.45 & 0.05 & 0.24 & \textbf{0.02} & \textbf{0.18} \\
\hline \hline
\multirow{5}*{CIFAR-100} & trial 1 & 7.99 & 17.45 & 7.99 & 17.22 & 8.22 & 15.35 \\
 & trial 2 & 7.94 & 17.72 & 7.97 & 16.78 & 8.07 & 16.74 \\
 & trial 3 & 7.94 & 17.57 & 8.02 & 16.86 & 8.15 & 15.73 \\
\cline{2-8}
 & Mean & 7.96 & 17.58 & 8.00 & 16.95 & \textbf{8.15} & \textbf{15.94} \\
\cline{2-8} 
 & Std & 0.03 & 0.13 & 0.03 & 0.24 & \textbf{0.08} & \textbf{0.72} \\

\hline
\end{tabular}
\end{center}
\label{table1}
\end{table}

\begin{figure}
\centering
\includegraphics[width=1.0\linewidth]{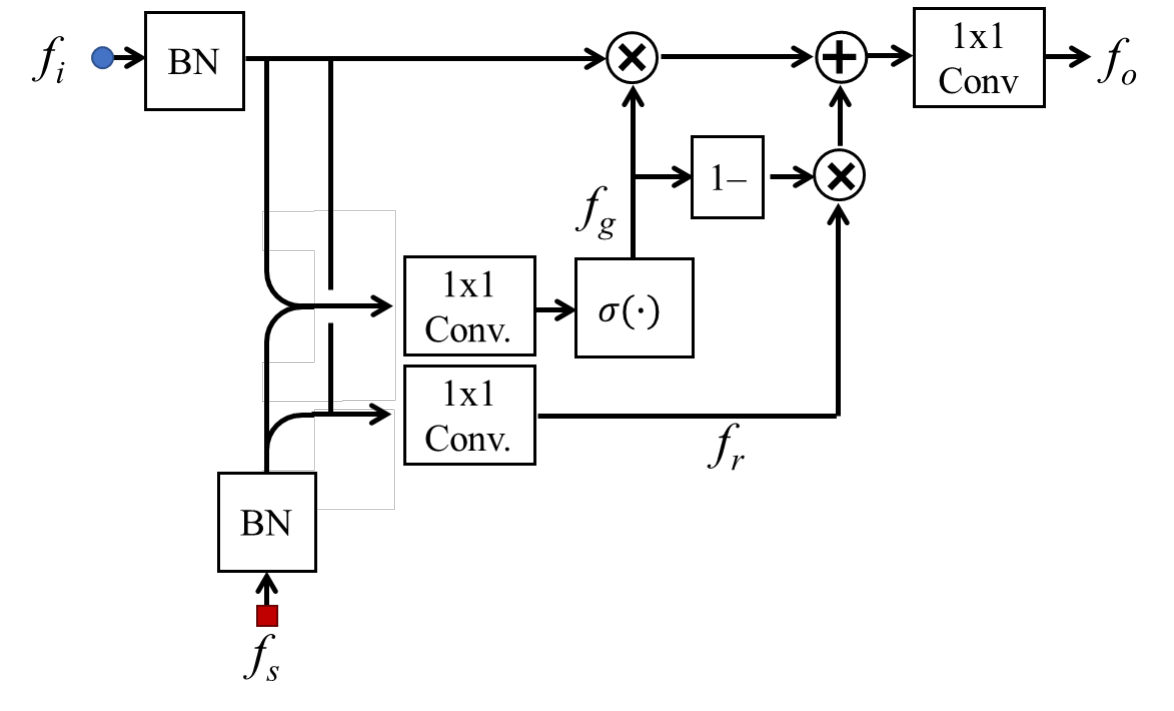}
\caption{Detailed architecture of the proposed GFFM. BN and Conv are the batch normalization~\cite{ioffe2015batch} and convolution operation, respectively. In addition, $\sigma(\cdot)$ indicates the sigmoid function. }
\label{fig:fig3}
\vspace{-0.5cm}
\end{figure}

To this end, inspired by the gated recurrent unit (GRU)~\cite{cho2014learning}, we design the gated feature fusion module (GFFM) that integrates features in both branches. In GRU, the two different features, \ie hidden and the current features, are combined by using the gating mechanism, which leads the fine performance in the field of natural language processing. Following this approach, we build the GFFM architecture that combines the features from the main and auxiliary branches using the gating mechanism. The architecture of GFFM is depicted in Fig.~\ref{fig:fig3}, where $f_m$ and $f_a$ indicate the features in the main and auxiliary branches, respectively. In the proposed method, since $f_m$ and $f_a$ have different levels of the features, we first utilize batch normalization (BN)~\cite{ioffe2015batch} to balance the scales of the features. Then, we blend the normalized features using the gating unit that controls the information flow through a gating mechanism. Specifically, the gating unit needs an input feature $f_i\in\mathbb{R}^{h\times w \times c_i}$ which will be refined, and requires a side feature $f_s\in\mathbb{R}^{h\times w \times c_s}$ that contains the information to be used for $f_i$ refinement. Here, $h$ and $w$ indicate the height and width of feature maps, respectively, whereas $c_\textrm{x}$ means the channel numbers of $f_\textrm{x}$. By using $f_i$ and $f_s$, the gate feature $f_g \in\mathbb{R}^{h\times w \times c_g}$, which controls the information passed on in the hierarchy is computed as follows:
\begin{equation}
    f_g = \sigma \big[\textrm{W}_g*(f_i \odot f_s)\big],
\label{eq3}
\end{equation}
where $*$ and $\odot$ indicate the convolution and channel-concatenation operations, respectively, whereas $\textrm{W}_g$ is a trainable weight in the convolution layer. $\sigma$ is a sigmoid function that produces output in the range [0, 1] to control how much to forget or keep the information. In addition, we generate the refinement feature $f_r \in\mathbb{R}^{h\times w \times c_r}$, which will be newly added to $f_i$, in a similar way. The $f_r$ is defined as 
\begin{equation}
    f_r = \big[\textrm{W}_f*(f_i \odot f_s)\big],
\label{eq4}
\end{equation}
where $\textrm{W}_f$ is a trainable weight in the convolution layer. By using these features, \ie $f_g$ and $f_r$, we produce the output feature $f_o \in \mathbb{R}^{h\times w \times c_o}$ as
\begin{equation}
    f_o = \textrm{W}_o*\big[f_g \otimes f_i + (1 - f_g) \otimes f_r\big],
\label{eq5}
\end{equation}
where $\otimes$ indicates element-wise multiplication and $\textrm{W}_o$ is a weight in the output convolution layer. Here, we could interpret the Eq.~\ref{eq5} as follows: the first term $f_g \otimes f_i$ means that the gating unit decides what it is relevant to keep in the $f_i$ and what information is unnecessary. In contrast, the second term $(1 - f_g) \otimes f_r$ indicates that $f_i$ is combined with new information coming from $f_s$. That means the network can effectively control information flow by adding necessary information derived from the earlier layer.

By using the proposed gating unit, the GFMM produces two different outputs $f^*_m$ and $f^*_a$ which are the refined main and side features, respectively. It is worth noting that the proposed method refines not only $f_m$ but also $f_a$ to store the updated features of the network; $f_a$ acts like a memory cell in the gated recurrent unit~\cite{cho2014learning}. When producing $f^*_m$, $f_m$ and $f_a$ are used for $f_i$ and $f_s$ of the gating unit, respectively. In contrast, when making $f^*_a$, $f_m$ and $f_a$ are used for $f_s$ and $f_a$ of the gating unit, respectively. Consequently, by adding the GFFM between the residual blocks, the generator produces the image while successfully considering the necessary features in the earlier layers.

\begin{table*}[t]
\caption{Network architectures of the discriminator (top) and generator (bottom) for each image resolution. The RB means the residual block presented in Fig.~\ref{fig:fig1}(d).}
\begin{center}
\begin{tabular}{c | c | c | c}
\hline
$32\times32$ resolution & $128\times128$ resolution & $256\times256$ resolution & $512\times512$ resolution \\
\hline
RGB image & RGB image & RGB image & RGB image \\
RB, down, 128 & RB, down, 64 & RB, down, 32 & RB, down, 16\\
RB, down, 128 & RB, down, 128 & RB, down, 64 & RB, down, 32\\
RB, 128 & RB, down, 256 & RB, down, 128 & RB, down, 64\\
RB, 128 & RB, down, 512 & RB, down, 256 & RB, down, 128\\
ReLU & RB, down, 512 & RB, down, 512 & RB, down, 256\\
Global sum pooling & RB, 512 & RB, down, 512 & RB, down, 512\\
Dense, 1 & ReLU & RB, 512 & RB, down, 512\\
 & Global sum pooling & ReLU & RB, 512\\
 & Dense, 1 & Global sum pooling & ReLU\\
 & & Dense, 1 & Global sum pooling\\
 & & & Dense, 1 \\
 
\hline
$32\times32$ resolution & $128\times128$ resolution & $256\times256$ resolution & $512\times512$ resolution \\
\hline
FC, $4 \times 4 \times 256$ & FC, $4 \times 4 \times 512$ & FC, $4 \times 4 \times 512$ & FC, $4 \times 4 \times 512$ \\
RB, up, 256 / GFFM, 256 & RB, up, 512 / GFFM, 512 & RB, up, 512 / GFFM, 512 & RB, up, 512 / GFFM, 512 \\
RB, up, 256 / GFFM, 256 & RB, up, 512 / GFFM, 512 & RB, up, 512 / GFFM, 512 & RB, up, 512 / GFFM, 512\\
RB, up, 256 / GFFM, 256 & RB, up, 256 / GFFM, 256 & RB, up, 256 / GFFM, 256 & RB, up, 256 / GFFM, 256\\
BN, ReLU & RB, up, 128 & RB, up, 128 / GFFM, 128  & RB, up, 128 / GFFM, 128\\
$3\times3$ conv, Tanh & RB, up, 64  / GFFM, 64 & RB, up, 64 / GFFM, 64 & RB, up, 64 / GFFM, 64\\
 & BN, ReLU & RB, up, 32 / GFFM, 32 & RB, up, 32 / GFFM, 32\\
& $3\times3$ conv, Tanh & BN, ReLU  & RB, up, 16 / GFFM, 16\\
& & $3\times3$ conv, Tanh  & BN, ReLU \\
& & & $3\times3$ conv, Tanh   \\

\hline
\end{tabular}
\end{center}
\label{tableGD}
\vspace{-0.5cm}
\end{table*}

\subsection{Implementation Details}
\label{subsec:3.2}
To reveal the superiority of the proposed method, we conducted extensive experiments using the various standard datasets that are widely used for measuring the GAN performance: CIFAR-10~\cite{krizhevsky2009learning}, CIFAR-100~\cite{krizhevsky2009learning}, LSUN~\cite{yu15lsun}, CelebA-HQ~\cite{karras2017progressive}, AFHQ~\cite{choi2020stargan}, and tiny-ImageNet~\cite{23deng2009imagenet, yao2015tiny} which is a subset of ImageNet~\cite{23deng2009imagenet}, consisting of the 200 selected classes. Specifically, among the various classes in the LSUN dataset, we employed the church and tower images in our experiments. The resolutions of the CIFAR-10 and CIFAR-100 datasets are $32\times 32$, whereas we resized the images in LSUN and tiny-ImageNet datasets to $128\times 128$ pixels. In addition, the CelebA-HQ and AFHQ datasets are used for training the networks that produce the $256\times 256$ and $512\times 512$ images. To train the network, we used the hinge-adversarial loss~\cite{lim2017geometric}. Since all network parameters in the discriminator and generator including the GFFM can be differentiated, we used the Adam optimizer~\cite{kingma2014adam} and set the user parameters, \ie $\beta _1$ and $\beta _2$, to 0 and 0.9, respectively.

For training the CIFAR-10 and CIFAR-100 datasets, we set the learning rate as 0.0002 for both the discriminator and generator. The discriminator was updated 5 times using different mini-batches while the generator is updated once. We set the batch size of the discriminator as 64 and trained the generator for 50k iterations. Following the previous papers~\cite{miyato2018cgans, miyato2018spectral, park2021GRB}, the generator was trained with a batch size twice as large as when training the discriminator. In other words, the generator and discriminator were trained with 128 and 64 batch sizes, respectively. On the other hand, for training the LSUN and tiny-ImageNet datasets, we employed a two-time scale update rule, called TTUR~\cite{heusel2017gans}, which set the learning rates of the generator and discriminator to 0.0001 and 0.0004, respectively. Since the TTUR method updates the discriminator a single time when the generator is updated once, we set both batch sizes of the discriminator and the generator to 32. The generator is updated to 300k iterations on the LSUN dataset and 1M iterations on the tiny-ImageNet dataset. In addition, for training the CelebA-HQ and AFHQ datasets, we employed the TTUR technique and set both batch sizes of the discriminator and the generator to 16. The generator is updated to 100k iterations. For all datasets, we decreased the learning rate linearly in the last 50,000 iterations. In addition, For training the CIFAR-10 and CIFAR-100 datasets the spectral normalization (SN)~\cite{miyato2018spectral} was only used for the discriminator, whereas the SN was applied to both generator and discriminator when training the LSUN, CelebA-HQ, AFHQ, and tiny-ImageNet datasets. 

In our experiments, we built generator and discriminator architectures following a strong baseline in~\cite{brock2018large}, called BigGAN. The detailed discriminator and generator architectures are described in Table~\ref{tableGD}. In the discriminator, the feature maps were down-sampled by utilizing the average pooling after the second convolution layer in the residual block. In the generator, the up-sampling (a nearest-neighbor interpolation) operation was located before the first convolution. In the GFFM, we set $c_g = c_r = c_o$, and $c_o$ in each module is described in Table~\ref{tableGD}. In addition, we matched the spatial size of $f_m$ and $f_a$ using the nearest neighbor interpolation technique after the BN of $f_a$. When $c_a$ and $c_o$ in the GFFM were different, we used $1 \times 1$ convolution having $c_o$ output channels before the BN of $f_a$ to match the number of channels. In the last GFFM, we only used $f^*_m$ to produce the image. On the other hand, to train the network in the cGAN framework, we replaced the BN in the generator with the conditional BN~\cite{dumoulin2017learned, brock2018large} and added the conditional projection layer in the discriminator~\cite{miyato2018cgans}. More detailed explanations are presented in~\cite{miyato2018cgans}.

\begin{table*}[t]
\caption{Comparison of the proposed method with conventional ones on the CIFAR-10 and CIFAR-100 datasets in terms of IS and FID. The best results are in bold.}
\begin{center}
\begin{tabular}{c | c | c | c | c | c | c | c || c | c | c | c | c | c }
\hline
\multicolumn{2}{c|}{} & \multicolumn{6}{c||}{GAN} & \multicolumn{6}{c}{cGAN} \\ 
\hline
\multicolumn{2}{c|}{Network} & \multicolumn{2}{c|}{SNGAN~\cite{miyato2018cgans}} & \multicolumn{2}{c|}{BigGAN~\cite{brock2018large}} & \multicolumn{2}{c||}{Proposed} & \multicolumn{2}{c|}{SNGAN~\cite{miyato2018cgans}} & \multicolumn{2}{c|}{BigGAN~\cite{brock2018large}} & \multicolumn{2}{c}{Proposed} \\
\hline
Dataset & & IS$\uparrow$ & FID$\downarrow$ & IS$\uparrow$ & FID$\downarrow$ & IS$\uparrow$ & FID$\downarrow$ & IS$\uparrow$ & FID$\downarrow$ & IS$\uparrow$ & FID$\downarrow$ & IS$\uparrow$ & FID$\downarrow$ \\
\hline
\multirow{5}*{CIFAR-10} & trial 1 & 7.79 & 13.81 & 7.70 & 13.98 & 8.06 & 10.92 & 8.06 & 10.26 & 8.09 & 9.36 & 8.27 & 7.96 \\
 & trial 2 & 7.76 & 13.29 & 7.80 & 13.69 & 8.11 & 10.96 & 8.02 & 10.37 & 7.97 & 9.62 & 8.20 & 8.20 \\
 & trial 3 & 7.76 & 13.29 & 7.87 & 13.09 & 8.12 & 11.12 & 7.99 & 10.01 & 8.02 & 9.38 & 8.32 & 7.83 \\
\cline{2-14}
 & Mean & 7.77 & 13.46 & 7.79 & 13.58 & \textbf{8.10} & \textbf{11.00}  & 8.03 & 10.21 & 8.03 & 9.45 & \textbf{8.26} & \textbf{8.00} \\
\cline{2-14} 
 & Std & 0.02 & 0.30 & 0.08 & 0.45 & \textbf{0.03} & \textbf{0.11} & 0.03 & 0.18 & 0.06 & 0.15 & \textbf{0.06} & \textbf{0.19} \\

\hline 
\multirow{5}*{CIFAR-100} & trial 1 & 8.08 & 17.76 & 7.99 & 17.45 & 8.30 & 14.56 & 8.62 & 14.62 & 8.88 & 13.17 & 9.34 & 10.39 \\
 & trial 2 & 8.08 & 17.93 & 7.94 & 17.72 & 8.20 & 15.04 & 8.62 & 14.39 & 8.89 & 13.20 & 9.37 & 10.17 \\
 & trial 3 & 8.05 & 17.55 & 7.94 & 17.57 & 8.35 & 14.87 & 8.80 & 14.85 & 8.87 & 13.01 & 9.28 & 10.25\\
\cline{2-14}
 & Mean & 8.07 & 17.75 & 7.96 & 17.58 & \textbf{8.29} & \textbf{14.82} & 8.68 & 14.62 & 8.88 & 13.13 & \textbf{9.33} & \textbf{ 10.27} \\
\cline{2-14} 
 & Std & 0.02 & 0.19 & 0.03 & 0.13 & \textbf{0.08} & \textbf{0.24} & 0.10 & 0.23 & 0.01 & 0.10 & \textbf{0.05} & \textbf{0.11}\\

\hline
\end{tabular}
\end{center}
\label{table2}
\end{table*}

\section{Experimental Results}
\label{sec4}
\subsection{Evaluation Metric}
\label{subsec:4.1}
FID~\cite{heusel2017gans} and IS~\cite{salimans2016improved} are the most widely used assessments that measure how well the generator synthesizes the image. More specifically, FID, which computes the Wasserstein distance between the distributions of the real and generated samples in the feature space of the Inception model, is defined as follows: 
\begin{equation}
    \textrm{F}(p,q) = \| \mu_p - \mu_q \|_2^2 + \mathrm{trace}(C_p +C_q - 2(C_p C_q)^{\frac{1}{2}}),
\end{equation}
where $ \{\mu_p,C_p \}$ and $\{\mu_q,C_q \}$ are the mean and covariance of the samples with distributions of real and generated images, respectively. A lower FID score indicates that the generated images have better quality. On the other hand, Salimans~\etal~\cite{salimans2016improved} demonstrated that IS is strongly correlated with the subjective human judgment of image quality. IS is expressed as 
\begin{equation}
I = \mathrm{exp}(E[D_{KL}(p(l|X)||p(l))]),
\end{equation}
where \textit{l} is the class label predicted by the Inception model~\cite{26szegedy2016rethinking} trained by using the ImageNet dataset~\cite{23deng2009imagenet}, and $p(l|X)$ and $p(l)$ represent the conditional class distributions and marginal class distributions, respectively. In contrast with the FID, the better the quality of the generated image, the higher the IS score. In our experiments, we randomly generated 50,000 samples and computed the FID and IS scores using the same number of real images.

\begin{table}[t]
\caption{Comparison of the proposed method with conventional ones on the CIFAR-10 and CIFAR-100 datasets in terms of IS and FID, when the number of network parameters is the same. BigGAN$^\dagger$ is a BigGAN network built with the dense block. }
\begin{center}
\begin{tabular}{c | c | c | c | c }
\hline
\multirow{3}*{Dataset} & \multirow{3}*{Metric} & BigGAN & BigGAN$^\dagger$ & Proposed \\
&  &  (5.68M & (5.68M & (5.68M\\
&  &  parameters) & parameters) & parameters)\\

\hline
\multirow{2}*{CIFAR-10} & IS$\uparrow$ & $7.82 \pm 0.03$ & $7.90 \pm 0.02$ & \textbf{8.10 $\pm$ 0.03}\\
\cline{2-5} 
& FID$\downarrow$ & $12.14 \pm 0.15$ & $11.66 \pm 0.23$ & \textbf{11.00 $\pm$ 0.11}\\
\hline
\multirow{2}*{CIFAR-100} & IS$\uparrow$ & $8.19 \pm 0.04$ & $8.16 \pm 0.02$ & \textbf{8.29 $\pm$ 0.08}\\
\cline{2-5} 
& FID$\downarrow$ & $16.15 \pm 0.16$ & $15.90 \pm 0.12$ & \textbf{14.82 $\pm$ 0.24}\\

\hline
\end{tabular}
\end{center}
\label{table3}
\end{table}

\begin{table}[t]
\caption{Performance evaluations on the CIFAR-10 and CIFAR-100 datasets in terms of IS and FID, when using the naive feature fusion modules.}
\begin{center}
\begin{tabular}{c | c | c | c | c }
\hline
\multirow{2}*{Dataset} & \multirow{2}*{Metric} & \multirow{2}*{Summation} &  \multirow{2}*{Concatenation} & GFFM \\
& & & & (Proposed)\\

\hline
\multirow{2}*{CIFAR-10} & IS$\uparrow$ & $7.99 \pm 0.08$ & $8.03 \pm 0.02$ & \textbf{8.10 $\pm$ 0.03}\\
\cline{2-5} 
& FID$\downarrow$ & $12.38 \pm 0.20$ & $11.59 \pm 0.17$ & \textbf{11.00 $\pm$ 0.11}\\
\hline
\multirow{2}*{CIFAR-100} & IS$\uparrow$ & $8.07 \pm 0.04$ & $8.18 \pm 0.01$ & \textbf{8.29 $\pm$ 0.08}\\
\cline{2-5} 
& FID$\downarrow$ & $16.98 \pm 0.10$ & $16.29 \pm 0.09$ & \textbf{14.82 $\pm$ 0.24}\\

\hline
\end{tabular}
\end{center}
\label{table_abl1}
\end{table}

\begin{table}[t]
\caption{Performance evaluations on the CIFAR-10 and CIFAR-100 datasets in terms of IS and FID to verify the effectiveness of the auxiliary branch.}
\begin{center}
\begin{tabular}{c | c | c | c | c }
\hline
\multirow{2}*{Dataset} & Shortcut & Auxiliary  &  \multirow{2}*{IS$\uparrow$} & \multirow{2}*{FID$\downarrow$}\\
& connection & branch & &\\
\hline
\multirow{4}*{CIFAR-10} & & & $7.76 \pm 0.12$ & $14.36 \pm 0.40$ \\
\cline{2-5} 
 & \checkmark & & $7.79 \pm 0.08$ & $13.58 \pm 0.45$ \\
 \cline{2-5} 
 & & \checkmark & $7.99 \pm 0.08$ & $11.20 \pm 0.11$ \\
 \cline{2-5} 
 & \checkmark & \checkmark & \textbf{8.10 $\pm$ 0.10} & \textbf{11.00 $\pm$ 0.11} \\
\hline
\multirow{4}*{CIFAR-100} & & & $7.88 \pm 0.06$ & $17.78 \pm 0.12$ \\
\cline{2-5} 
 & \checkmark & & $7.96 \pm 0.03$ & $17.58 \pm 0.13$ \\
 \cline{2-5} 
 & & \checkmark & \textbf{8.30 $\pm$ 0.09} & $15.17 \pm 0.32$ \\
 \cline{2-5} 
  & \checkmark & \checkmark & 8.29 $\pm$ 0.08 & \textbf{14.82 $\pm$ 0.24} \\
\hline

\end{tabular}
\end{center}
\label{table_abl2}
\end{table}

\subsection{Experimental Results}
\label{subsec:4.3}
To reveal the effectiveness of the proposed method, we conducted preliminary experiments on the image generation task using the CIFAR-10 and CIFAR-100 datasets. In our experiments, we trained the network three times from scratch to show that the performance improvement is not due to the lucky weight initialization. Table~\ref{table2} summarizes the comprehensive results that compare the GAN performance between the proposed and conventional methods. On both CIFAR-10 and CIFAR-100 datasets, the proposed method consistently outperformed the current state-of-the-art methods, \ie SNGAN and BigGAN, in terms of FID and IS scores. These results indicate that the proposed method is more effective than the existing residual block-based generators. In addition, in the cGAN framework, the proposed method showed remarkable performance on both datasets and exhibited better performance than its counterpart by a large margin. For instance, on the CIFAR-100 dataset, the proposed method achieved the FID of 10.27, which was about 29.75\% and 21.78\% better than SNGAN and BigGAN, respectively. Based on these results, we believe that the proposed method can achieve outstanding performance in both GAN and GAN schemes.

\begin{table}[t]
\caption{Comparison of the proposed method with conventional ones on the LSUN-church and LSUN-tower datasets in terms of FID. The best results are in bold.}
\begin{center}
\begin{tabular}{c | c | c | c | c}
\hline
\multicolumn{2}{c|}{Network} & SNGAN & BigGAN & Proposed \\
\hline
Dataset & & FID$\downarrow$ & FID$\downarrow$ & FID$\downarrow$ \\
\hline
 & trial 1 & 7.83 & 8.18 & 5.59 \\
 & trial 2 & 8.16 & 7.88 & 5.56 \\
LSUN- & trial 3 & 8.22 &  8.11 & 5.40 \\
\cline{2-5}
church & Mean & 8.07 & 8.06 & \textbf{5.51} \\
\cline{2-5} 
 & Std & 0.21 & 0.15 & \textbf{0.10} \\
\hline 
 & trial 1 & 12.42 & 13.89 & 8.77 \\
 & trial 2 & 12.29 & 12.38 & 9.90 \\
LSUN- & trial 3 & 12.54 & 11.98 & 8.85 \\
\cline{2-5}
tower & Mean & 12.42 & 12.78 & \textbf{9.17} \\
\cline{2-5} 
 & Std & 0.12 & 0.99 & \textbf{0.63} \\
\hline
\end{tabular}
\end{center}
\label{table4}
\end{table}

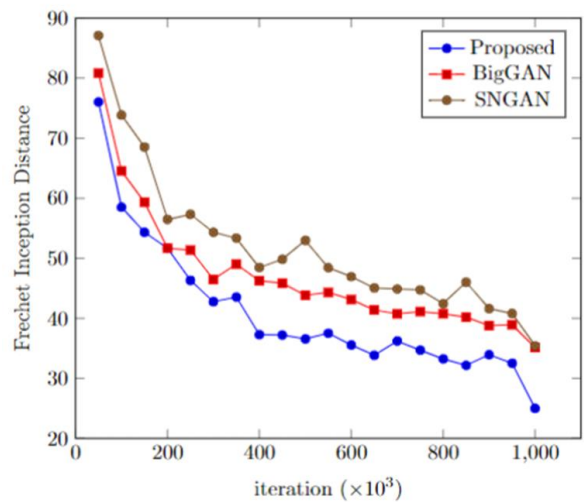
\begin{figure}
\centering
    \begin{tikzpicture}[scale=0.9]
        \begin{axis}[ ymax = 90, ymin = 20, xmin = -0.2,
                xlabel=iteration $(\times 10^3)$,
                ylabel=Frechet Inception Distance,
                legend pos=north east]
            \addplot+[error bars/.cd,
                       y dir=both, y explicit]
                    coordinates {
                    (50, 76.03) 
                    (100, 58.53) 
                    (150, 54.33) 
                    (200, 51.67)
                    (250, 46.32)
                    (300, 42.77)
                    (350, 43.53)
                    (400, 37.27)
                    (450, 37.20)
                    (500, 36.56)
                    (550, 37.50)
                    (600, 35.55)
                    (650, 33.83)
                    (700, 36.21)
                    (750, 34.68)
                    (800, 33.23)
                    (850, 32.16)
                    (900, 33.94)
                    (950, 32.5)
                    (1000, 25.00)
                    };
                \addlegendentry{Proposed}

            \addplot+[error bars/.cd,
                       y dir=both, y explicit]
                    coordinates {
                    (50, 80.85) 
                    (100, 64.54) 
                    (150, 59.32) 
                    (200, 51.67)
                    (250, 51.37)
                    (300, 46.47)
                    (350, 49.01)
                    (400, 46.25)
                    (450, 45.83)
                    (500, 43.84)
                    (550, 44.31)
                    (600, 43.12)
                    (650, 41.39)
                    (700, 40.73)
                    (750, 41.11)
                    (800, 40.75)
                    (850, 40.19)
                    (900, 38.77)
                    (950, 38.91)
                    (1000, 35.13)
                    };
                \addlegendentry{BigGAN}
                
                            \addplot+[error bars/.cd,
                       y dir=both, y explicit]
                    coordinates {
                    (50, 87.08) 
                    (100, 73.86) 
                    (150, 68.53) 
                    (200, 56.44)
                    (250, 57.32)
                    (300, 54.32)
                    (350, 53.35)
                    (400, 48.45)
                    (450, 49.82)
                    (500, 52.96)
                    (550, 48.42)
                    (600, 46.94)
                    (650, 45.06)
                    (700, 44.89)
                    (750, 44.72)
                    (800, 42.42)
                    (850, 46.00)
                    (900, 41.62)
                    (950, 40.81)
                    (1000, 35.42)
                    };
                \addlegendentry{SNGAN}

        \end{axis}
    \end{tikzpicture}
    \vspace{-0.4cm}
    \caption{Comparison of FID scores over training iterations. Blue, red, and yellow lines indicate the FID scores of the proposed method, BigGAN, and SNGAN, respectively.}
\label{fig:fig_fid}
\end{figure}

On the other hand, one might perceive that the performance improvement is due to the increased number of network parameters. We agree that the proposed method requires additional network parameters used in the GFFM. To alleviate this concern, we conducted ablation studies that match the network parameters of the conventional methods with that of the proposed one; we increased the channel of the residual block to match the number of network parameters. Since the proposed method follows BigGAN as the baseline model, we compared the performance of the proposed method with that of BigGAN. As presented in Table~\ref{table3}, even using the same number of network parameters, the proposed method exhibited better performance than the conventional methods. Moreover, since we have already shown that the dense block is more effective than the residual block in Section~\ref{sec2.2}, we also measured the performance of BigGAN built with the dense block (BigGAN$^\dagger$ in Table~\ref{table3}). In our experiments, we adjusted the channel of the dense block to match the number of network parameters. As described in Table~\ref{table3}, the dense block-based BigGAN exhibited better performance than the residual block-based BigGAN but still showed weaker performance than the proposed method. These results contain two important meanings: First, the performance improvement is not only due to the increased number of network parameters. Second, compared with the residual and dense blocks, the proposed method is more effective in improving the GAN performance by propagating the information in the earlier layers to the later ones successfully.

\begin{figure}
\centering
    \begin{tikzpicture}[scale=0.9]
        \begin{axis}[ ymax = 27, ymin = 10, xmin = -0.2,
                xlabel=iteration $(\times 10^3)$,
                ylabel=Inception Score,
                legend pos=south east]
            \addplot+[error bars/.cd,
                       y dir=both, y explicit]
                    coordinates {
                    (50, 11.12) 
                    (100, 13.54) 
                    (150, 14.69) 
                    (200, 15.02)
                    (250, 16.88)
                    (300, 17.39)
                    (350, 17.06)
                    (400, 18.70)
                    (450, 19.07)
                    (500, 19.54)
                    (550, 19.21)
                    (600, 19.94)
                    (650, 19.75)
                    (700, 20.44)
                    (750, 20.44)
                    (800, 21.68)
                    (850, 21.58)
                    (900, 20.66)
                    (950, 21.63)
                    (1000, 25.57)
                    };
                \addlegendentry{Proposed}

            \addplot+[error bars/.cd,
                       y dir=both, y explicit]
                    coordinates {
                    (50, 10.47) 
                    (100, 11.87) 
                    (150, 13.41) 
                    (200, 14.69)
                    (250, 14.50)
                    (300, 15.60)
                    (350, 15.08)
                    (400, 16.17)
                    (450, 16.48)
                    (500, 16.22)
                    (550, 16.92)
                    (600, 16.87)
                    (650, 17.24)
                    (700, 17.57)
                    (750, 17.66)
                    (800, 17.90)
                    (850, 17.90)
                    (900, 17.83)
                    (950, 18.22)
                    (1000, 20.23)
                    };
                \addlegendentry{BigGAN}
                
            \addplot+[error bars/.cd,
                   y dir=both, y explicit]
                    coordinates {
                    (50, 9.64) 
                    (100, 11.24) 
                    (150, 11.84) 
                    (200, 13.81)
                    (250, 13.71)
                    (300, 14.16)
                    (350, 14.55)
                    (400, 15.85)
                    (450, 15.40)
                    (500, 15.16)
                    (550, 15.81)
                    (600, 16.26)
                    (650, 16.92)
                    (700, 17.27)
                    (750, 16.88)
                    (800, 17.90)
                    (850, 16.96)
                    (900, 18.23)
                    (950, 17.83)
                    (1000, 20.52)
                    };
            \addlegendentry{SNGAN}

        \end{axis}
    \end{tikzpicture}
    \vspace{-0.4cm}
    \caption{Comparison of IS scores over training iterations. Blue, red, and yellow lines indicate the IS scores of the proposed method, BigGAN, and SNGAN, respectively.}
\label{fig:fig_is}
\vspace{-0.4cm}
\end{figure}
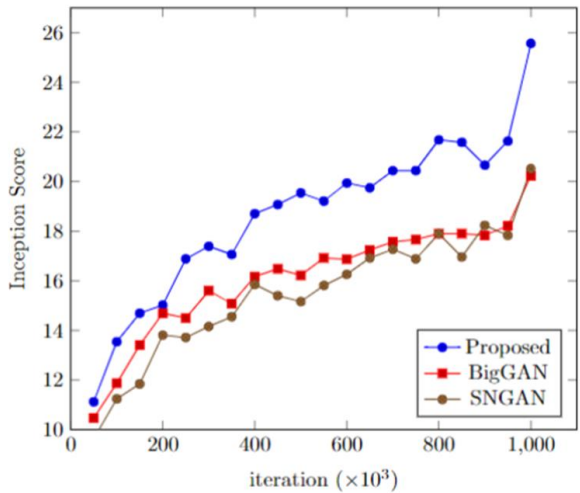

\begin{table}[t]
\caption{Comparison of the final FID and IS scores on the tiny-ImageNet dataset. The best results are in bold.}
\begin{center}
\begin{tabular}{c | c | c | c | c }
\hline
Dataset & Metric & SNGAN & BigGAN & Proposed \\
\hline
tiny- & IS$\uparrow$ & 20.52 & 20.23 & \textbf{25.57}\\
\cline{2-5}
ImageNet & FID$\downarrow$ & 35.42 & 35.13 & \textbf{25.00} \\

\hline
\end{tabular}
\end{center}
\label{table:table5}
\vspace{-0.3cm}
\end{table}

Moreover, we conducted ablation studies to reveal the ability of the auxiliary branch with GFFM. First, to show the effectiveness of the GFFM, we measured the GAN performance with prevalent feature fusion methods, \ie feature summation and concatenation. The results in Table~\ref{table_abl1} show that the GFFM achieves better performance on the CIFAR-10 and CIFAR-100 datasets. In addition, we observed that even using the feature summation or concatenation techniques, the generator with the auxiliary branch outperforms conventional methods. To prove the ability of the auxiliary branch more clearly, we conducted other ablation studies that how much the auxiliary branch improves the GAN performance. In fact, the proposed method contains two factors that assist the information flow: a shortcut connection in the residual block~\cite{he2016deep} and the auxiliary branch. To prove the effectiveness of each part, we evaluated the GAN performance of the proposed method, while omitting the shortcut connection or auxiliary branch. As summarized in Table~\ref{table_abl2}, the generator with the auxiliary branch accomplishes better performance than that with the shortcut connection. Furthermore, we observed that the generator achieves the lowest FID score by using the shortcut connection and auxiliary branch together. Based on these results of the ablation studies, we concluded that the GFFM and auxiliary branch improves the GAN performance effectively. 

\begin{figure}
\centering
\includegraphics[width=0.9\linewidth]{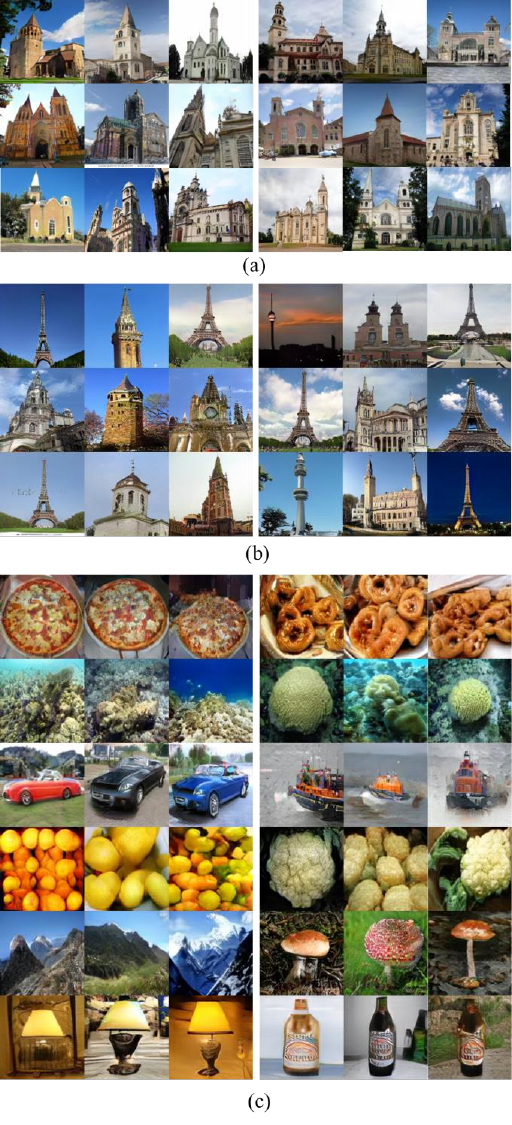}
\caption{Samples of the generated images. (a) Generated images on the LSUN-church dataset, (b) Generated images on the LSUN-tower dataset, (c) Generated images on the tiny-ImageNet dataset.}
\label{fig_gen}
\vspace{-0.5cm}
\end{figure}

To show the effectiveness of the proposed method when synthesizing images on challenging datasets, we trained the network using the LSUN-church and LSUN-tower datasets. Specifically, on the LSUN-church and -tower datasets, the network is trained by following the experimental setting for GAN. As shown in Table~\ref{table4}, the proposed method exhibited a notable performance compared to the conventional methods by a large margin. That means the proposed method is also effective to generate the image with the complex scene. Moreover, we trained the network using the tiny-ImageNet dataset, following the experimental setting for cGAN. Since we already proved that the performance improvement is not due to the lucky weight initialization in Tables~\ref{table2},~\ref{table3} and~\ref{table4}, we trained the network a single time from scratch on the tiny-ImageNet dataset. Instead, to show the superiority of the proposed method more reliably, we drew a graph representing the FID and IS scores over training iterations. As shown in Figs.~\ref{fig:fig_fid} and~\ref{fig:fig_is}, the proposed method consistently outperformed the conventional methods, \ie SNGAN and BigGAN, during the training procedure. The final FID and IS scores are summarized in Table~\ref{table:table5}. Fig.~\ref{fig_gen} shows the samples of the generated images. This results show that proposed method is effective in synthesizing images with complex scenes. Based on these results, we concluded that the proposed method can generate visually pleasing images on challenging datasets.

\begin{figure}
\centering
\includegraphics[width=0.95\linewidth]{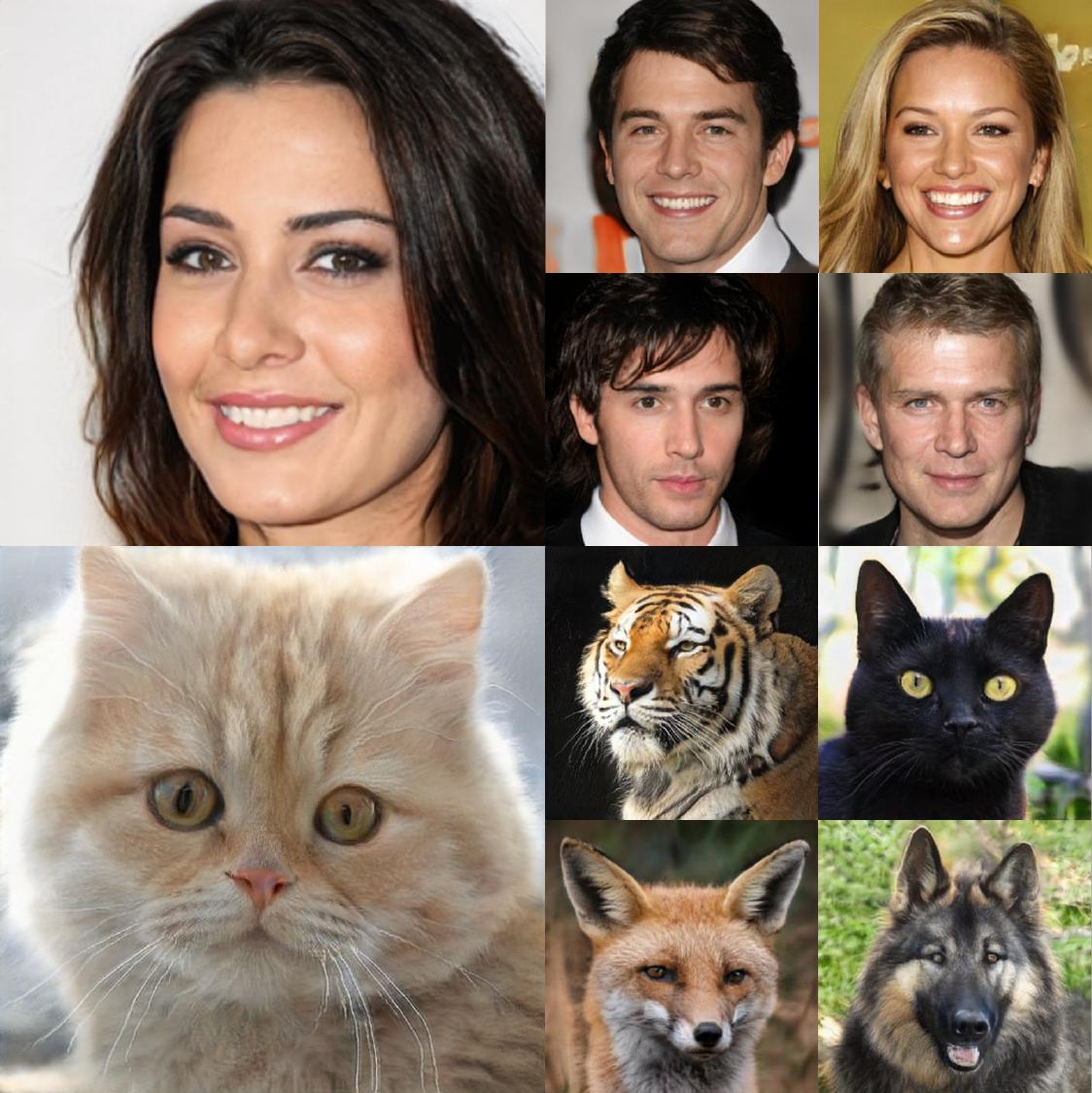}
\caption{Generated images with $256\times256$ and $512\times512$ resolutions on the Celeb-HQ and AFHQ datasets.}
\label{fig:CA-HQ}
\vspace{-0.5cm}
\end{figure}

\begin{table}[t]
\caption{Comparison of the proposed method with conventional ones on the CelebA-HQ and AFHQ datasets in terms of FID.}
\begin{center}
\begin{tabular}{c | c | c | c || c | c}
\hline
Dataset & Size & BigGAN & Proposed & StyleGAN2 & Proposed \\
\hline
CelebA & 256 & 16.27 & \textbf{11.25} & 9.91 & \textbf{7.90} \\
\cline{2-6}
HQ & 512 & 27.40 & \textbf{23.43} & 9.18 & \textbf{7.74}\\
\hline 
 \multirow{2}*{AFHQ} & 256 & 19.02 & \textbf{16.35} & 10.12 & \textbf{9.13}\\
 \cline{2-6}
 & 512 & 22.47 & \textbf{17.40} & 10.73 & \textbf{8.86}\\
\hline
\end{tabular}
\end{center}
\label{table:table_CAHQ}
\vspace{-0.5cm}
\end{table}

Furthermore, to show the ability of the proposed method for generating high-resolution images, we performed additional experiments using the CelebA-HQ~\cite{karras2017progressive} and AFHQ~\cite{choi2020stargan} datasets. In this brief, we trained the networks to produce $256\times256$ and $512\times512$ images. However, since the CelebA-HQ and AFHQ datasets are composed of a small number of images, it often suffers from the overfitting problem of the discriminator. To alleviate this problem, we employed PConv to the discriminator~\cite{park2021generative}. The results in Table~\ref{table:table_CAHQ} show that the proposed method is also effective to generate high-resolution images. To prove the generalization ability of the proposed method, we conducted additional experiments by setting StyleGAN2~\cite{karras2020analyzing} as the baseline model; we replaced the residual block in StyleGAN2 with the proposed AB block. In our experiments, we set the same number of batch sizes and the training iterations with the BigGAN-based experiments. As reported in Table~\ref{table:table_CAHQ}, the proposed method shows lower FID scores than StyleGAN2. In addition, as depicted in Fig.~\ref{fig:CA-HQ}, the proposed method with StyleGAN2 produces visually plausible images. Based on these results, we expect that the proposed method can be utilized to generate high-quality images with various baseline models. Indeed, this study does not intend to design an optimal generator and discriminator architecture specialized for the auxiliary branch and GFFM. There might be another network architecture that improves the GAN performance and produces more high-quality images. Instead, this brief focuses on verifying whether it is possible to achieve better GAN performance by simply adding the auxiliary branch and GFFM. 

\section{Conclusion}
\label{sec5}
This brief has introduced a novel generator architecture with the auxiliary branch and has shown the remarkable performance of the proposed method in various experiments. In addition, we have proposed the GFFM that effectively blends the features in the main and auxiliary branches. The main advantage of the proposed method is that it significantly boosts the GAN performance with a slight number of additional network parameters. Furthermore, this brief has demonstrated the generalization ability of the proposed method in various aspects through high-resolution image generation and extensive experiments. Therefore, we expect that the proposed method is applicable to various GAN-based image generation techniques. However, this brief has aimed at introducing a novel generator architecture specialized to the GAN. We agree that the proposed method works well for the GAN-based image generation technique, but might cause some issues in other networks used for different applications such as image-to-image translation. There might be another network architecture that shows more general ability than the proposed method. Although this brief only has covered the effectiveness of the proposed method in the field of GAN-based image generation technique, we have shown that the proposed method boosts the GAN performance significantly. In our future work, we plan to further investigate a novel generator architecture that could cover various applications.

\bibliographystyle{IEEEtran}

\bibliography{egbib.bib}

\end{document}